\documentclass{article}

\usepackage{neurips_2021}

\usepackage[utf8]{inputenc} % allow utf-8 input
\usepackage[T1]{fontenc}    % use 8-bit T1 fonts
\usepackage{hyperref}       % hyperlinks
\usepackage{url}            % simple URL typesetting
\usepackage{booktabs}       % professional-quality tables
\usepackage{amsfonts}       % blackboard math symbols
\usepackage{nicefrac}       % compact symbols for 1/2, etc.
\usepackage{microtype}      % microtypography
\usepackage{xcolor}         % colors
\usepackage{graphicx}

\usepackage{caption}
\usepackage{algorithm}
\usepackage{algorithmic}
\usepackage{enumitem}
\usepackage{thmtools}
\usepackage{amsmath}
\usepackage{amssymb}
\usepackage{cleveref}

\newtheorem{theorem}{Theorem}[section]
\newtheorem{lemma}[theorem]{Lemma}

\newtheorem{proposition}[theorem]{Proposition}
\newtheorem{corollary}[theorem]{Corollary}

\newtheorem{definition}{Definition}[section]

\newcommand{\qed}{\hfill $\square$}

\title{Factored couplings in multi-marginal optimal transport via difference of convex programming}

\author{
  Quang Huy Tran \\
  Univ. Bretagne-Sud, CNRS, IRISA \\
  F-56000 Vannes \\
  \texttt{quang-huy.tran@univ-ubs.fr} \\
  \And
  Hicham Janati \\
  École Polytechnique, CMAP, UMR 7641 \\
  F-91120 Palaiseau \\
  \texttt{hicham.janati@polytechnique.edu} \\
  \AND
  Ievgen Redko \\
  Univ Lyon, UJM-Saint-Etienne, CNRS, UMR 5516 \\
  F-42023 Saint-Etienne \\
  \texttt{ievgen.redko@univ-st-etienne.fr} \\
  \And
  Rémi Flamary \\
  École Polytechnique, CMAP, UMR 7641 \\
  F-91120 Palaiseau \\
  \texttt{remi.flamary@polytechnique.edu} \\
  \And
  Nicolas Courty \\
  Univ. Bretagne-Sud, CNRS, IRISA \\
  F-56000 Vannes \\
  \texttt{nicolas.courty@irisa.fr} \\
}

\begin{document}

\maketitle

\begin{abstract}
    Optimal transport (OT) theory underlies many emerging machine learning (ML) methods nowadays solving a wide range of tasks such as generative modeling, transfer learning and information retrieval. These latter works, however, usually 
    build upon a traditional OT setup with two distributions, while leaving a more general multi-marginal OT formulation somewhat 
    unexplored. In this paper, we study the multi-marginal OT (MMOT) problem and unify several popular OT methods under its umbrella 
    by promoting structural information on the coupling. We show that incorporating such structural 
    information into MMOT results in an instance of a difference of convex (DC) programming problem allowing us to solve it numerically. 
    Despite high computational cost of the latter procedure, the solutions provided by DC optimization are usually as qualitative as 
    those obtained using currently employed optimization schemes.
    %We show that it interpolates between MMOT and factorized MMOT. Then we present the approximation scheme and its accelerated version. We illustrate in some experiments.
\end{abstract}

%%%%%%%%%%%%%%%%%%%%%%%%%%%%%%%%%%%%%%%%%%%%%%
\section{Introduction}
%%%%%%%%%%%%%%%%%%%%%%%%%%%%%%%%%%%%%%%%%%%%%%
Broadly speaking, the classic OT problem provides a principled approach for transporting one probability distribution onto another 
following the principle of the least effort. Such a problem, and the distance on the space of probability distributions derived from it, 
arise in many areas of machine learning (ML) including generative modeling, transfer learning and information retrieval, where OT has 
been successfully applied. A natural extension of classic OT, in which the admissible transport plan (a.k.a coupling) can have more 
than two prescribed marginal distributions, is called the multi-marginal optimal transport (MMOT) \citep{Gangbo98}. 
The latter has several attractive properties: it enjoys a duality theory \citep{Kellerer84} and finds connections with the 
probabilistic graphical models \citep{Haasler20} and the Wasserstein barycenter problem \citep{Agueh11} used for data averaging. 
While being less popular than the classic OT with two marginals, MMOT is a very useful framework on its own with some notable recent 
applications in generative adversarial networks \citep{Cao19}, clustering \citep{Mi21} and domain adaptation 
\citep{HuiLCHY18,HeZKSC19}, to name a few. 

The recent success of OT in ML is often attributed to the entropic regularization \citep{Cuturi13} where the authors imposed a 
constraint on the coupling matrix forcing it to be closer to the independent coupling given by the rank-one product of the marginals. 
Such a constraint leads to the appearance of the strongly convex entropy term in the objective function and allows the entropic 
OT problem to be solved efficiently using simple Sinkhorn-Knopp matrix balancing algorithm. In addition to this, it was also noticed 
that structural constraints on the coupling and cost matrices allow to reduce the high computational cost and sample complexity 
of the classic OT problem \citep{Genevay19,Forrow18,Chiheng21,Meyer21a}. However, none of these works considered a much more 
challenging case of doing so in a multi-marginal setting. On the other hand, while the work of \citep{Haasler20} considers the MMOT 
problem in which the cost tensor induced by a graphical structure, it does not naturally promote the factorizability of 
transportation plans.
%Additionally, several OT formulations dealing with calculating OT for measures supported on points in incomporable spaces were also proposed. Finally, due to the high sample complexity of solving OT in high-dimensions, several works also sought to promote structural low-rank constraints on the coupling \citep{Forrow18,Chiheng21}, which has been recently shown to be closely related to the low-nonnegative rank coupling \citep{Meyer21a}. 

\paragraph{Contributions} In this paper, we define and study a general MMOT problem with structural penalization on the coupling matrix. 
We start by showing that a such formulation includes several popular OT methods as special cases and allows to gain deeper insights 
into them. We further consider a relaxed problem where the hard constraint is replaced by a regularization term and show that it leads 
to an instance of the difference of convex programming problem. A numerical study of the solutions obtained when solving the latter 
in cases of interest highlights their competitive performance when compared to solutions provided by the optimization 
strategies used previously. 
%We propose a relaxation of MMOT called \textit{MMOT-DC}, whose name is motivated by the formulation of MMOT 
%problem and the connection with DC programming. After showing some preliminary properties, we show that MMOT-DC interpolates 
%between factorized MMOT and MMOT. We then derive the DC algorithm to solve MMOT-DC.

%\paragraph{Factorized transportation plan in OT.} The idea of factored coupling is first studied by  \citep{Forrow18}, which is recently shown to be closely related to the low nonnegative rank coupling \citep{Meyer21a}.  A variation of factored coupling is also proposed in \citep{Chiheng21}. Similar counterparts in MMOT are the well-known third lower bound  of Gromov-Wasserstein distance \citep{Memoli07,Memoli11} and the recent CO-Optimal transport (COOT) \citep{Redko20}, where both can be recast as factorized MMOT problem in which the admissible transportation plan can be written as product measure. In this paper, we study a relaxed version of such problem.

\section{Preliminary knowledge}

\paragraph{Notations.} For each integer $n \geq 1$, we write $[n] := \{1,...,n\}$. For any discrete probability measure 
$\mu$ with finite support, its negative entropy is defined as $H(\mu) = \langle \mu, \log \mu \rangle$, where the logarithm 
operator is element-wise, with the convention that $0 \log 0 = 0$. Here, $\langle \cdot, \cdot \rangle$ denotes the Frobenius 
inner product. The Kullback-Leibler divergence between two discrete probability measures $\mu$ and $\nu$ with finite supports is
defined as 
\begin{equation*}
  \text{KL}(\mu \vert \nu) = 
  \begin{cases}
    \langle \mu, \log\frac{\mu}{\nu} \rangle, \text{ if } \mu \text{ is absolutely continuous with respect to } \nu \\
    \infty, \text{ otherwise}.
  \end{cases}
\end{equation*}
where the division operator in the logarithm is element-wise.

In what follows, given an integer $N \geq 1$, for any positive integers $a_1,..., a_N$, we call 
$P \in \mathbb R^{a_1 \times ... \times a_N}$ a $N$-D tensor. In particular, a $1$-D tensor is a vector and $2$-D tensor is a matrix. 
A tensor is a probability tensor if its entries are nonnegative and the sum of all entries is $1$. 
Given $N$ probability vectors $\mu_1, ..., \mu_N$, we write $\mu = (\mu_n)_{n=1}^N$.
We denote $\Sigma$ the set of $N$-D probability tensors and $U(\mu) \subset \Sigma$ the set of nonnegative tensors whose $N$ 
marginal distributions are $\mu_1, ..., \mu_N$. In this case, any coupling in $U(\mu)$ is said to be \textit{admissible}.

\paragraph{Multi-marginal OT problem.} Given a collection of $N$ probability vectors $\mu = (\mu_n \in \mathbb R^{a_n})_{n=1}^N$ 
and a $N$-D cost tensor $C \in \mathbb R^{a_1 \times ... \times a_N}$, the MMOT problem reads
\begin{equation*}
  \text{MMOT}(\mu) = \inf_{P \in U(\mu)} \langle C, P \rangle.
\end{equation*}
In practice, such a formulation is intractable to optimize in a discrete setting as it results in a linear program where the number 
of constraints grows exponentially in $N$. A more tractable strategy for solving MMOT is to consider the following entropic 
regularization problem
\begin{equation} \label{MMOT_primal}
  \inf_{P \in U(\mu)} \langle C, P \rangle + \varepsilon H(P).
\end{equation}
which can be solved using Sinkhorn's algorithm \citep{Benamou14}. We refer the interested reader to Supplementary materials for algorithmic details.

%\iffalse
%\paragraph{Difference of convex programming.} A popular class of non-convex optimization studies the problem of the following form.
%\begin{equation*}
%  \min_{x \in C} f(x) - g(x),
%\end{equation*}
%where $C$ is convex set, $f$ and $g$ are convex, proper and lower semi-continuous functions. The objective function is known as a difference of convex (DC) function. Such problem can be solved by the DC algorithm \citep{Tao86,Tao97}. Its idea is simple: 
%one constructs two sequences $(x_k)_k$ and $(y_k)_k$ and at the iteration $k$, the function $g$ is replaced by its affine minorant 
%$g_k(x) := g(x_k) + \langle x - x_k, y_k \rangle$, with $y_k \in \partial g(x_k)$, where $\partial g(x)$ is the sub-differential of 
%$g$ at $x$. Then, we minimise the resulting convex function. This procedure is closely related to the maximisation minimisation 
%algorithm \citep{Ortega70}, where the surrogate function is $f - g_k$.
%\fi

%%%%%%%%%%%%%%%%%%%%%%%%%%%%%%%%%%%%%%%%%%%%
\section{Factored Multi-marginal Optimal Transport}
%%%%%%%%%%%%%%%%%%%%%%%%%%%%%%%%%%%%%%%%%%%%
In this section, we first define a factored MMOT (F-MMOT) problem where we seek to promote a structure on the optimal coupling 
given such as a factorization into a tensor product. Interestingly, such a formulation can be shown to include several other 
OT problems as special cases. Then, we introduce a relaxed version called MMOT-DC where the factorization constraint is 
smoothly promoted through a Kullback-Leibler penalty.

\subsection{Motivation} 

Before a formal statement of our problem, we first give a couple of motivating examples showing why and when structural constraints on the coupling matrix can be beneficial. To this end, first note that a trivial example of the usefulness of such constraints in OT is the famous entropic regularization. Indeed, while most of the works define the latter by adding negative entropy of the coupling to the classic OT objective function directly, the original idea was to constraint the sought coupling to remain close (to some extent) to a rank-one product of the two marginal distributions. The appearance of negative entropy in the final objective function is then only a byproduct of such constraint due to the decomposition of the KL divergence into a sum of three terms with two of them being constant. Below we give two more examples of real-world applications related to MMOT problem where a certain decomposition imposed on the coupling tensor can be desirable. 
\paragraph{Multi-source multi-target translation.} A popular task in computer vision is to match images across different domains in order to perform the so-called image translation. Such tasks are often tackled within the GAN framework where one source domain from which the translation is performed, is matched with multiple target domains modeled using generators. While MMOT was applied in this context by \citep{Cao19} when only one source was considered, its application in a multi-source setting may benefit from structural constraints on the coupling tensor incorporating the human prior on what target domains each source domain should be matched to. 
\paragraph{Multi-task reinforcement learning.} In this application, the goal is to learn individual policies for a set of agents while taking into account the similarities between them and hoping that the latter will improve the individual policies. A common approach is to consider an objective function consisting of two terms where the first term is concerned with learning individual policies, while the second forces a consensus between them. Similar to the example considered above, MMOT problem was used to promote the consensus across different agents' policies in \citep{Cohen21}, even though such a consensus could have benefited from a prior regarding the semantic relationships between the learned tasks. 
%
%We can cite other examples motivating the introduction of structural constraints in MMOT problem but the bottom line of it remains unchanged: imposing a certain structure on the optimal coupling tensor is a way of incorporating the human-based priors to the MMOT problem that reflect the domain knowledge about the problem at hand. We now proceed to the formal introduction of this idea. 

\subsection{Factored MMOT and its relaxation}
We start by giving several definitions used in the following parts of the paper. 
%We call $P \in \mathbb R^{a_1 \times ... \times a_N}$ a $N$-D tensor. When $N=1$, we simply call it a vector and when $N=2$, 
%it is a matrix. 
\begin{definition}[Tuple partition]
 Given two integers $N \geq M \geq 2$, a sequence of tuples $\mathcal T = (\mathcal T_m)_{m=1}^M$, is called a 
 \underline{tuple partition} of the $N$-tuple $(1,...,N)$ if the tuples $\mathcal T_1, ..., \mathcal T_M$ are nonempty and disjoint, 
 and their concatenation in this order gives $(1,...,N)$. 
\end{definition}
Here, we implicitly take into account the order of the tuple, which is not the case for the partition of the set $[N]$. If 
there exists a tuple in $\mathcal T$ which contains only one element, then we say $\mathcal T$ is \textit{degenerate}.

\begin{definition}[Marginal tensor]
  Given a tensor $P \in \mathbb R^{a_1 \times ... \times a_N}$ and a tuple partition $\mathcal T = (\mathcal T_m)_{m=1}^M$, 
  we call $P_{\# \mathcal T_m}$ its \underline{$\mathcal T_m$-marginal tensor}, by summing $P$ over all dimensions not in $\mathcal T_m$. 
  We write $P_{\# \mathcal T} = P_{\# \mathcal T_1} \otimes ... \otimes P_{\# \mathcal T_M} \in \mathbb R^{a_1 \times ... \times a_N}$ 
  the tensor product of its marginal tensors. 
\end{definition}
For example, for $M=N=2$, we have $\mathcal T_1 = (1)$ and $\mathcal T_2 = (2)$. So, given a matrix 
$P \in \mathbb R^{a_1 \times a_2}$, its marginal tensors $P_{\# \mathcal T_1}$ and $P_{\# \mathcal T_2}$ are simply vectors in 
$\mathbb R^{a_1}$ and $\mathbb R^{a_2}$, respectively, defined by $(P_{\# \mathcal T_1})_i = \sum_j P_{ij}$ and 
$(P_{\# \mathcal T_2})_j = \sum_i P_{ij}$ for $(i,j) \in [a_1] \times [a_2]$. The tensor product 
$P_{\# \mathcal T} \in \mathbb R^{a_1 \times a_2}$ is then defined by 
$(P_{\#\mathcal T})_{ij} = (P_{\# \mathcal T_1})_i (P_{\# \mathcal T_2})_j$.

Clearly, if $P$ is a probability tensor, then so are its marginal tensors and tensor product.

Suppose $\mathcal T_m = (p,...,q)$ for some $m \in [M]$ and $1 \leq p \leq q \leq N$. We denote 
$\Sigma_{\mathcal T_m}$ the set of probability tensors in $\mathbb R^{a_p \times ... \times a_q}$ and 
$U_{\mathcal T_m} \subset \Sigma_{\mathcal T_m}$ the set 
of probability tensors in $\mathbb R^{a_p \times ... \times a_q}$ whose $(r)$-marginal vector is $\mu_r$, for every $r = p,...,q$.

%%%%%%%%%%%%%%%%%%%%%%%%%%%%%%%
\begin{definition}[Factored MMOT]
  Given a collection of histograms $\mu = (\mu_n)_{n=1}^N$ and a tuple partition $\mathcal T = (\mathcal T_m)_{m=1}^M$, 
  we consider the following OT problem
  \begin{equation} \label{factor_mmot}
    \text{F-MMOT}( \mathcal T, \mu) = \inf_{P \in U_{\mathcal T}} \langle C, P \rangle,
  \end{equation}
  where $U_{\mathcal T} \subset U(\mu)$ is the set of admissible couplings which can be factorized as a tensor product of $M$ 
  component probability tensors in $\Sigma_{\mathcal T_1}, ..., \Sigma_{\mathcal T_M}$. 
\end{definition}
Several remarks are in order here. First, one should note that the partition considered above is in general not degenerate meaning 
that the decomposition can involve tensors of an arbitrary order $<N$. Second, the decomposition in this setting depicts the prior 
knowledge regarding the tuples of measures which should be independent: the couplings for the measures from different tuples will 
be degenerate and the optimal coupling tensor will be reconstructed from couplings of each tuple separately.  
Third, suppose the partition $(\mathcal T_m)_{m=1}^M$ is not degenerate and $M=2$, i.e. the tensor is factorized as product of 
two tensors, the problem \ref{factor_mmot} is equivalent to a variation of low nonnegative rank OT problem (see Appendix for a proof).

As for the existence of the solution to this problem, we have that $U_{\mathcal T}$ is compact because it is a close subset of the 
compact set $U(\mu)$, which implies that the problem \ref{factor_mmot} always admits a solution. Furthermore, observe that
\begin{equation*}
  \begin{split}
    U_{\mathcal T} &= \{ P \in U(\mu): P = P_1 \otimes ... \otimes P_M, \text{where } P_m \in \Sigma_{\mathcal T_m}, \forall m = 1,...,M \} \\
    &= \{ P \in \Sigma: P = P_1 \otimes ... \otimes P_M, \text{where } P_m \in U_{\mathcal T_m}, \forall m = 1,...,M \}.
  \end{split}
\end{equation*}
Thus, the problem F-MMOT can be rewritten as
\begin{equation*}
  \text{F-MMOT}( \mathcal T, \mu) = \inf_{\substack{P_m \in U_{\mathcal T_m} \\ \forall m = 1,...,M}} 
  \langle C, P_1 \otimes ... \otimes P_M \rangle.
\end{equation*}
So, if $\mathcal T_1,...,\mathcal T_M$ are $2$-tuples and two marginal distributions corresponding to each $U_{\mathcal T_m}$ are 
identical and uniform, then by Birkhoff's theorem \citep{Birkhoff46}, the problem \ref{factor_mmot} admits an optimal solution in 
which each component tensor $P_m$ is a permutation matrix. 

\paragraph{Two special cases.} When $N = 4$ and $M=2$ with $\mathcal T_1 = (1,2)$ and $\mathcal T_2 = (3,4)$, the problem 
\ref{factor_mmot} becomes the CO-Optimal transport (COOT) \citep{Redko20}, where the two component tensors are known as 
\textit{sample} and \textit{feature} couplings. If furthermore, $a_1 = a_3, a_2=a_4$, and $\mu_1 = \mu_3, \mu_2=\mu_4$, it becomes a 
lower bound of the discrete Gromov-Wasserstein (GW) distance \citep{Memoli11}. This means that our formulation can be seen as a 
generalization of several OT formulations.

%%%%%%%%%%%%%%%%%%%%%%%%%%%%%%%%%%%%%%%%%%%%%
Observe that if a probability tensor $P$ can be factorized as a tensor product of probability tensors, i.e. 
$P = P_1 \otimes ... \otimes P_M$, then each $P_m$ is also the $\mathcal T_m$-marginal tensor of $P$. In this case, 
we have $P = P_{\# \mathcal T}$. This prompts us to consider the following relaxation of factored MMOT, where the hard constraint 
$U_{\mathcal T}$ is replaced by a regularization term.
\begin{definition}[Relaxed Factored MMOT]
  Given $\varepsilon \geq 0$, a collection of measures $\mu$ and a tuple partition $\mathcal T$, 
  we define the following problem:
  \begin{equation} \label{relax_mmot}
    \text{MMOT-DC}_{\varepsilon}( \mathcal T, \mu) = 
    \inf_{P \in U(\mu)} \langle C, P \rangle + \varepsilon \text{KL}(P \vert P_{\# \mathcal T}).
  \end{equation}
\end{definition}
From the exposition above, one can guess that this relaxation is reminiscent of the entropic regularization in MMOT and 
coincides with it when $M = N$. As such, it also recovers the classical entropic OT. One should note that the choice of the KL 
divergence is not arbitrary and its advantage will become clear when it comes to the algorithm. %A well known
A special case of the problem \ref{relax_mmot} is when $M = N$, we recover the entropic-regularized MMOT problem, up to a constant.

After having defined the two optimization problems, we now set on exploring their theoretical properties. 

\subsection{Theoretical properties}
%The following properties are direct consequences of the definition.
Intuitively, the relaxed problem is expected to allow for solutions with a lower value of the final objective function. We formally prove the validity of this intuition below.
%%%%%%%%%%%%%%%%%%%%%%%%%%%%%%%%%%%%%%%%%%%%%
\begin{proposition}[Preliminary properties] \label{MMOT_dc_prop}
  Given a collection of histograms $\mu$ and a tuple partition $\mathcal T$,
  \begin{enumerate}
    \item For every $\varepsilon \geq 0$, we have $\text{MMOT}(\mu) \leq 
    \text{MMOT-DC}_{\varepsilon}(\mathcal T, \mu) \leq \text{F-MMOT}( \mathcal T, \mu)$.
    \item For every $\varepsilon > 0, \text{MMOT-DC}_{\varepsilon}( \mathcal T, \mu ) = 0$ if and only if 
    $\text{F-MMOT} (\mathcal T, \mu) = 0$.
  \end{enumerate}
\end{proposition}
%%%%%%%%%%%%%%%%%%%%%%%%%%%%%%%%%%%%%%%%%%%%%
An interesting property of MMOT-DC is that it interpolates between MMOT and F-MMOT. Informally, 
for very large $\varepsilon$, the KL divergence term dominates, so the optimal transport plans tend to be factorizable. 
On the other hand, for very small $\varepsilon$, the KL divergence term becomes negligible and we approach MMOT. 
The result below formalizes this intuition.
%%%%%%%%%%%%%%%%%%%%%%%%%%%%%%%%%%%%%%%%%%%%%
\begin{proposition}[Interpolation between MMOT and F-MMOT] \label{interpolation_prop}
  For any tuple partition $\mathcal T$ and for $\varepsilon > 0$, 
  let $P_{\varepsilon}$ be a minimiser of the problem $\text{MMOT-DC}_{\varepsilon}(\mathcal T, \mu)$.
  \begin{enumerate}
    \item When $\varepsilon \to \infty$, one has $\text{MMOT-DC}_{\varepsilon}(\mathcal T, \mu) \to 
    \text{F-MMOT}(\mathcal T, \mu)$. In this case, any cluster point of the sequence of minimisers 
    $(P_{\varepsilon})_{\varepsilon}$ is a minimiser of $\text{F-MMOT}(\mathcal T, \mu)$.

    \item When $\varepsilon \to 0$, then $\text{MMOT-DC}_{\varepsilon}(\mathcal T, \mu) \to \text{MMOT}(\mu)$. 
    In this case, any cluster point of the sequence of minimisers $(P_{\varepsilon})_{\varepsilon}$ is a minimiser of 
    $\text{MMOT}(\mu)$.
  \end{enumerate}
\end{proposition}
%%%%%%%%%%%%%%%%%%%%%%%%%%%%%%%%%%%%%%%%%%%%%
\paragraph{GW distance revisited.} Somewhat surprisingly, the relaxation \ref{relax_mmot} also allows us to prove the equality 
between GW distance and COOT in the discrete setting. Let $\mathcal X$ be a 
finite subset (of size $m$) of a certain metric space. Denote $C_x \in \mathbb R^{m \times m}$ its similarity matrix (e.g. distance 
matrix). We define similarly the set $\mathcal Y$ of size $n$ and the corresponding similarity matrix $C_y \in \mathbb R^{n \times n}$. 
We also assign two discrete probability measures $\mu_x \in \mathbb R^m$ and $\mu_y \in \mathbb R^n$ to $\mathcal X$ and $\mathcal Y$, 
respectively. The GW distance is then defined as
\begin{equation*}
  \text{GW}(C_x, C_y) = \inf_{Q \in U(\mu_x, \mu_y)} \langle L(C_x, C_y), Q \otimes Q \rangle,
\end{equation*}
and the COOT reads
\begin{equation*}
  \text{COOT}(C_x, C_y) = \inf_{\substack{Q_s \in U(\mu_x, \mu_y) \\ Q_f \in U(\mu_x, \mu_y)}} 
  \langle L(C_x, C_y), Q_s \otimes Q_f \rangle,
\end{equation*}
where $L(C_x,C_y) \in \mathbb R^{m \times n \times m \times n}$ represents the $4$-D cost tensor induced by the matrices $C_x$ and $C_y$, 
and $U(\mu, \nu)$ is the set of couplings in $\mathbb R^{m \times n}_{\geq 0}$ whose two marginal distributions are $\mu$ and
$\nu$. When $C_x$ and $C_y$ are two squared Euclidean distance matrices, and $L(C_x,C_y)$ is of the form 
$\big(L(C_x,C_y)\big)_{i,j,k,l} = \vert (C_x)_{i,k} - (C_y)_{j,l} \vert^2$, it can be shown that the GW distance is equal 
to the COOT \citep{Redko20}. This is also true when $L(C_x, C_y)$ is a negative definite kernel \citep{Sejourne20}. 
Here, we establish a weaker case where this equality still holds.
%%%%%%%%%%%%%%%%%%%%%%%%%%%%%%%%%%%%%%%%%%%%%
\begin{corollary} \label{kernel_gw_coot}
  If $L(C_x, C_y)$ defines a conditionally negative definite kernel on $(\mathcal X \times \mathcal Y)^2$, then we have the equality 
  between GW distance and COOT. Furthermore, if $(Q_s^*,Q_f^*)$ is a solution of the COOT problem, then $Q_s^*$ and $Q_f^*$ are 
  two solutions of the GW problem. In particular, when $L(C_x, C_y)$ induces a strictly positive definite kernel 
  $\exp \big( -\frac{L(C_x, C_y)}{\varepsilon} \big)$, for every $\varepsilon > 0$, we have $Q_s^* = Q_f^*$.
\end{corollary}
%%%%%%%%%%%%%%%%%%%%%%%%%%%%%%%%%%%%%%%%%%%%%
The proof relies on the connection between MMOT-DC and COOT shown in the proposition \ref{interpolation_prop}, and given a 
$4$-D solution of MMOT-DC, we can construct another $4$-D solutions whose $\mathcal T_1$ and $\mathcal T_2$-marginal matrices are 
identical, under the assumption of the cost tensor. The proof of the second claim is deferred to the Appendix.

\section{Numerical solution} \label{sec:algo}
%%%%%%%%%%%%%%%%%%%%%%%%%%%%%%%%%%%%%%%%%%%%%
We now turn to the computational aspect of the problem \ref{relax_mmot}. First, note that for any tuple partition 
$\mathcal T = (\mathcal T_m)_{m=1}^M$ and probability tensor $P$, the KL divergence term can be decomposed as 
\begin{equation*}
  \text{KL}(P \vert P_{\# \mathcal T}) = H(P) - \sum_{m=1}^m H_m(P),
\end{equation*}
where the function $H_m$ defined by $H_m(P) := H(P_{\# \mathcal T_m})$ is continuous and convex with respect to $P$. 
Now, the problem \ref{relax_mmot} becomes
\begin{equation} \label{relax}
  \text{MMOT-DC}_{\varepsilon}(\mathcal T, \mu) = \inf_{P \in U(\mu)} 
  \langle C, P \rangle + \varepsilon H(P) - \varepsilon \sum_{m=1}^M H_m(P).
\end{equation}
This is nothing but a Difference of Convex (DC) programming problem (which explains the name MMOT-DC), 
thanks to the convexity of the set $U(\mu)$ and the entropy function $H$. Thus, it can be solved by the DC algorithm 
\citep{Tao86,Tao97} as follows: at the iteration $t$,
\begin{enumerate}
  \item Calculate $G^{(t)} \in \partial(\sum_{m=1}^M H_m)(P^{(t)})$.
  \item Solve $P^{(t+1)} \in \arg\min_{P \in U(\mu)} \langle C -
  \varepsilon G^{(t)}, P \rangle + \varepsilon H(P)$.
\end{enumerate}
%%%%%%%%%%%%%%%%%%%%%%%%%%%%%%%%%%%%%%%%%%%%%%
This algorithm is very easy to implement. Indeed, the second step is an entropic-regularized MMOT problem, which admits a unique 
solution, thanks to the strict convexity of the objective function. Such solution can be found by the Sinkhorn algorithm 
\ref{algo:dual_mmot}. In the first step, the gradient can be calculated explicitly. 
For the sake of simplicity, we illustrate the calculation in a simple case, where $M=2$ and $N=4$ with 
$\mathcal T_1$ and $\mathcal T_2$ are two $2$-tuples. The function $H_1 + H_2$ is continuous, so 
$G^{(t)} = \nabla_P (H_1 + H_2)(P^{(t)})$. Given a $4$-D probability tensor $P$, we have
\begin{equation*}
  H_1(P) + H_2(P) = \sum_{i,j,k,l} P_{i,j,k,l} 
  \log\big( \sum_{i,j} P_{i,j,k,l} \big) + P_{i,j,k,l} \log\big( \sum_{k,l} P_{i,j,k,l} \big).
\end{equation*}
So,
\begin{equation*} \label{optim_condition}
  \frac{\partial (H_1 + H_2)}{\partial P_{i,j,k,l}} = \log \left( \sum_{i,j} P_{i,j,k,l} \right) + 
  \frac{P_{i,j,k,l}}{\sum_{i,j} P_{i,j,k,l}} + 
  \log \left( \sum_{k,l} P_{i,j,k,l} \right) + \frac{P_{i,j,k,l}}{\sum_{k,l} P_{i,j,k,l}}.
\end{equation*}
The complete DC algorithm for the problem \ref{relax} can be found in the algorithm \ref{algo:dc_MMOT}.
%%%%%%%%%%%%%%%%%%%%%%%%%%%%%%%%%%%%%%%%%%%%%%
\begin{algorithm}[!t]
  \caption{DC algorithm for the problem \ref{relax_mmot}.}
  \textbf{Input.} Cost tensor $C$, tuple partition $(\mathcal T_m)_{m=1}^M$, collection of histograms $\mu = (\mu_n)_{n=1}^N$, 
  hyperparameter $\varepsilon > 0$, initialization $P^{(0)}$, tuple of initial dual vectors for the 
  Sinkhorn step $(f_1^{(0)},...,f_N^{(0)})$.

  \textbf{Output.} Tensor $P \in U(\mu)$.

  While not converge
  \begin{enumerate}
    \item Gradient step: compute the gradient of the convex term $G^{(t)} = \sum\limits_{m=1}^M \nabla_P H_m(P^{(t)})$.
    \item Sinkhorn step: solve 
    \begin{equation*}
      P^{(t+1)} = \arg\min_{P \in U(\mu)} \langle C - \varepsilon G^{(t)}, P \rangle + \varepsilon H(P),
    \end{equation*}
    using the Sinkhorn algorithm \ref{algo:dual_mmot}, with the tuple of initial dual vectors $(f_1^{(0)},...,f_N^{(0)})$.
  \end{enumerate}
  \label{algo:dc_MMOT}
\end{algorithm}
%%%%%%%%%%%%%%%%%%%%%%%%%%%%%%%%%%%%%%%%%%%%%
We observed that initialization is crucial to the convergence of algorithm, which
is not surprising for a non-convex problem. To accelerate the algorithm for large $\varepsilon$, 
we propose to use the warm-start strategy, which is similar to the one used in the entropic OT problem with very 
small regularization parameter \citep{Schmitzer19}. Its idea is simple: we consider an increasing finite sequence 
$(\varepsilon_n)_{n=0}^N$ approaching $\varepsilon$ such that the solution $P_{\varepsilon_0}$ of the problem 
$\text{MMOT-DC}_{\varepsilon_0}(\mathcal T, \mu)$ can be estimated quickly and accurately using the initialization $P^{(0)}$. Then we solve each 
successive problem $\text{MMOT-DC}_{\varepsilon_n}(\mathcal T, \mu)$ using the previous solution $P_{\varepsilon_{n-1}}$ as initialization. Finally, 
the problem $\text{MMOT-DC}_{\varepsilon}(\mathcal T, \mu)$ is solved using the solution $P_{\varepsilon_N}$ as initialization.
%%%%%%%%%%%%%%%%%%%%%%%%%%%%%%%%%%%%%%%%%%%%%
\begin{algorithm}[H]
  \caption{DC algorithm with warm start for the problem \ref{relax_mmot}.}
  \textbf{Input.} Cost tensor $C$, tuple partition $\mathcal T = (\mathcal T_m)_{m=1}^M$, collection of histograms $\mu = (\mu_n)_{n=1}^N$, 
  hyperparameter $\varepsilon > 0$, initialization $P^{(0)}$, initial $\varepsilon_0 > 0$, step size $s > 1$, 
  tuple of initial dual vectors $(f_1^{(0)}, ..., f_N^{(0)})$.

  \textbf{Output.} Tensor $P \in U(\mu)$.
  \begin{enumerate}
    \item While $\varepsilon_0 < \varepsilon$:
    \begin{enumerate}
      \item Using algorithm \ref{algo:dc_MMOT}, solve the problem $\text{MMOT-DC}_{\varepsilon_0}(\mathcal T, \mu)$ with initialization $P^{(0)}$ and 
      $(f_1^{(0)}, ..., f_N^{(0)})$ to find the solution $P_{\varepsilon_0}$ and its associated tuple of dual vectors 
      $(f_1^{(\varepsilon_0)}, ..., f_N^{(\varepsilon_0)})$.
      \item Set $P^{(0)} = P_{\varepsilon_0}, f_i^{(0)} = f_i^{(\varepsilon_0)}$, for $i=1,...,N$.
      \item Increase regularization: $\varepsilon_0 := s \varepsilon_0$.
    \end{enumerate}
    
    \item Using algorithm \ref{algo:dc_MMOT}, solve the problem $\text{MMOT-DC}_{\varepsilon}(\mathcal T, \mu)$ 
    using the initialization $P^{(0)}$ and $(f_1^{(0)},..., f_N^{(0)})$.
  \end{enumerate}
  \label{algo:acc_dc_MMOT}
\end{algorithm}
%%%%%%%%%%%%%%%%%%%%%%%%%%%%%%%%%%%%%%%%%%%%%
% \begin{algorithm}[H]
%   \caption{First accelerated DC algorithm for the problem \ref{relax_mmot}}
%   \textbf{Input.} Matrices $X \in \mathbb R^{n_x \times d_x}$ and $Y \in \mathbb R^{n_y \times d_y}$, discrete probability measures 
%   $\mu_{n_x}, \mu_{n_y}, \mu_{d_x}$ and $\mu_{d_y}$, hyperparameter $\varepsilon > 0$, initialization $P^{(0)}$, $Q^{(0)} = P^{(0)}$, 
%   integer $q \geq 1$.

%   \textbf{Output.} Tensor $P \in \mathbb R^{n_x \times n_y \times d_x \times d_y}$.
%   \begin{enumerate}
%     \item Compute $a_{t+1} = \frac{1 + \sqrt{1 + 4 a_t^2}}{2}$.
%     \item Compute $Q^{(t)} = P^{(t)} + \frac{a_t-1}{a_{t+1}} (P^{(t)} - P^{(t-1)})$.
%     \item Denote $F$ the objective function in \ref{relax_mmot}. If
%     \begin{equation*}
%       \min_{ijkl} Q^{(t)}_{ijkl} > 0 \;\;\; \text{ and } \;\;\; F(Q^{(t)}) \leq \max_{i=(0,t-q),...,t} F(P^{(i)})  
%     \end{equation*}
%     then set $h^{(t)} = \nabla (H_n + H_d)(Q^{(t)})$. Otherwise, set $h^{(t)} = \nabla (H_n + H_d)(P^{(t)})$.
    
%     \item Solve 
%     \begin{equation*}
%       P^{(t+1)} \in \arg\min_{P \in U_{xy}} \langle L(X,Y) - \varepsilon h^{(t)}, P \rangle + \varepsilon H(P)
%     \end{equation*}
%     using the Sinkhorn algorithm \ref{algo:dual_mmot}.
%   \end{enumerate}
%   \label{acc_dc_MMOT_v1}
% \end{algorithm}

%%%%%%%%%%%%%%%%%%%%%%%%%%%%%%%%%%%%%%%%%%%%%%
\section{Experimental evaluation} \label{sec:exp}
%%%%%%%%%%%%%%%%%%%%%%%%%%%%%%%%%%%%%%%%%%%%%%
In this section, we illustrate the use of MMOT-DC on simulated data. Rather than performing experiments in full generality, 
we choose the setting where $N = 4$ and $M=2$ with $\mathcal T_1 = (1,2)$ and $\mathcal T_2 = (3,4)$, 
so that we can compare MMOT-DC with other popular solvers of COOT and GW distance. Given two matrices $X$ and $Y$, we always consider the $4$-D cost tensor $C$, 
where $C_{i,j,k,l} = \vert X_{i,k} - Y_{j,l} \vert^2$. On the other hand, we are not interested in the $4$-D minimiser of MMOT-DC, 
but only in its two $\mathcal T_1, \mathcal T_2$-marginal matrices.

\paragraph{Solving COOT on a toy example.} We generate a random matrix $X \in \mathbb R^{30 \times 25}$, whose entries are drawn independently 
from the uniform distribution on the interval $[0,1)$. We equip the rows and columns of $X$ with two discrete uniform distributions 
on $[30]$ and $[25]$. We fix two permutation matrices $Q_s \in \mathbb R^{30 \times 30}$ (called sample permutation) and 
$Q_f \in \mathbb R^{25 \times 25}$ (called feature permutation), then calculate $Y = Q_s X Q_f$. We also equip the rows and columns of $Y$ 
with two discrete uniform distributions on $[30]$ and $[25]$. 

\begin{figure}[t]
  \centering
  \includegraphics[width=0.8\textwidth,height=0.8\textheight,keepaspectratio]{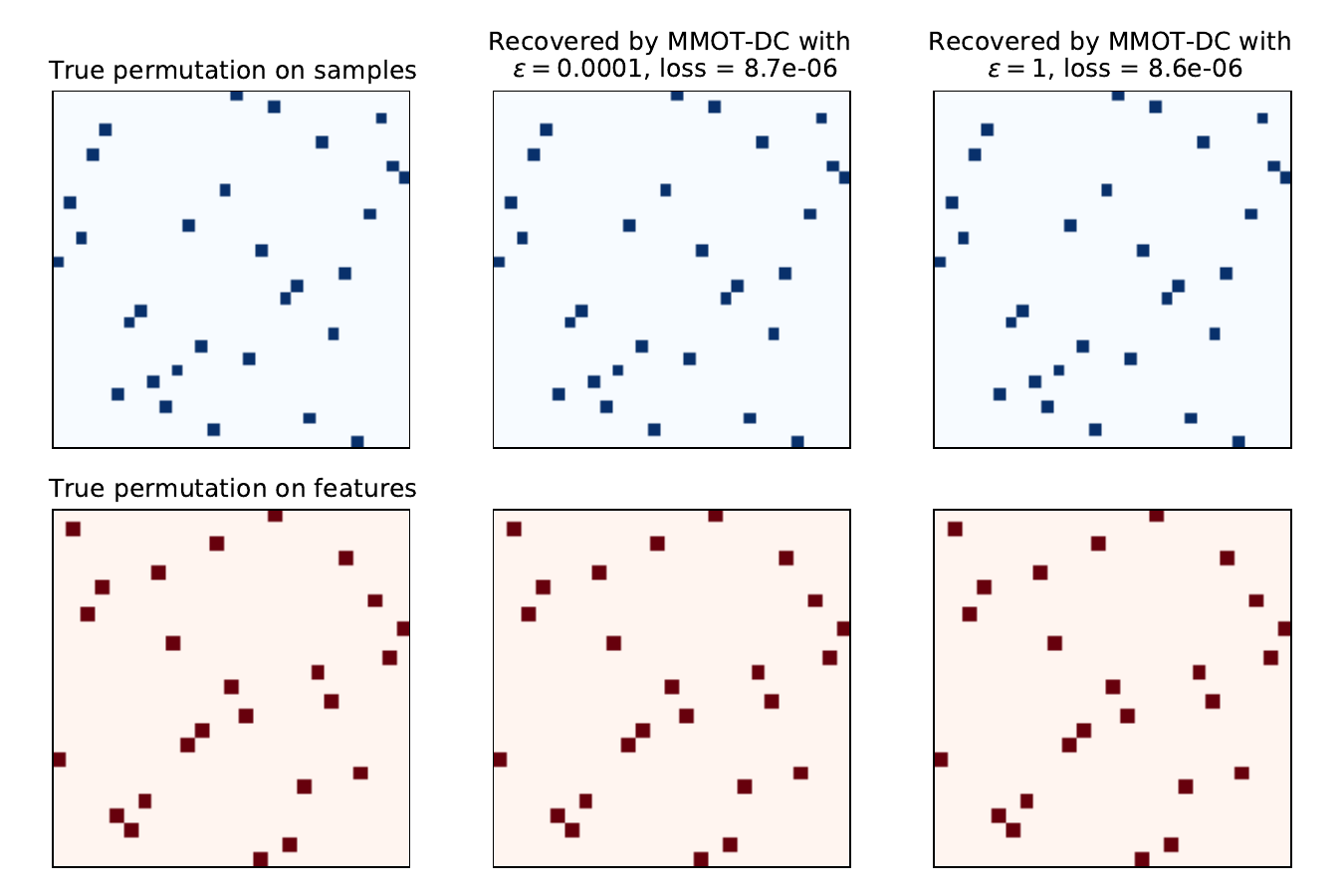}
  \caption{Couplings generated by COOT and MMOT-DC on the matrix recovering task.}
  \label{fig:permu}
\end{figure}

It is not difficult to see that $\text{COOT}(X,Y) = 0$ because $(Q_s, Q_f)$ is a solution. As COOT is a special case of F-MMOT, 
we see that $\text{MMOT-DC}_{\varepsilon}(\mathcal T, \mu) = 0$, for every $\varepsilon > 0$, 
by proposition \ref{MMOT_dc_prop}. In this experiment, we will check if marginalizing the minimizer of MMOT-DC allows us to recover 
the permutation matrices $Q_s$ and $Q_f$.
As can be seen from the figure \ref{fig:permu}, MMOT-DC can recover the permutation positions, for various 
values of $\varepsilon$. On the other hand, it can not recover the true sparse permutation matrices because the Sinkhorn algorithm 
applied to the MMOT problem implicitly results in a dense tensor, thus having dense marginal matrices. For this reason, 
the loss only remains very close to zero, but never exactly.
%%%%%%%%%%%%%%%%%%%%%%%%%%%%%%%%%%%%%%%
%%%%%%%%%%%%%%%%%%%%%%%%%%%%%%%%%%%%%%%

We also plot, with some abuse of notation, the histograms of the difference between
the $(1,3), (1,4), (2,3), (2,4)$-marginal matrices of MMOT-DC and their corresponding counterparts from F-MMOT. 
In this example, in theory, as the optimal tensor $P$ of F-MMOT can be factorized as 
$P = P_{\# \mathcal T_1} \otimes P_{\# \mathcal T_2} = Q_s \otimes Q_f$, 
it is immediate to see that $P_{\# (1,3)} = P_{\# (1,4)} = P_{\# (2,3)} = P_{\# (2,4)} \in \mathbb R^{30 \times 25}$ 
are uniform matrices whose entries are $\frac{1}{750}$.
%%%%%%%%%%%%%%%%%%%%%%%%%%%%%%%%%%%%%%%
\begin{figure}[t]
  \centering
  \includegraphics[width=1.\textwidth,height=1.\textheight,keepaspectratio]{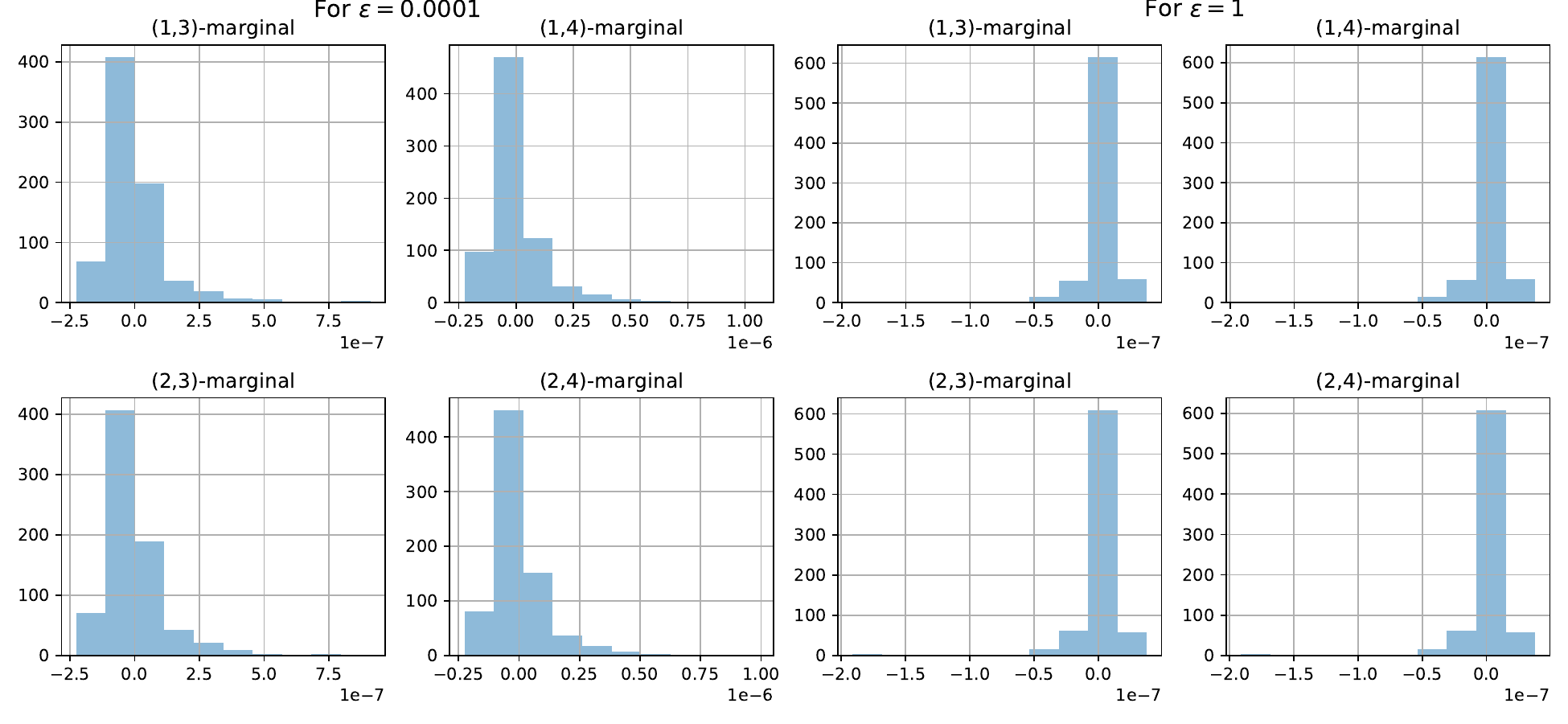}
  \caption{Histograms of difference between true independent marginal matrices and their approximations. We see that the marginal matrices obtained 
  by the algorithm \ref{algo:dc_MMOT} approximate well the theoretical uniform matrices.}
  \label{fig:other_marg}
\end{figure}
\paragraph{Quality of the MMOT-DC solutions. \label{expe:2}} 

% In the following example, our evaluation metric is the COOT loss $\langle C, P \otimes Q \rangle$, 
% where the smaller the loss, the better.

Now, we consider the situation where the true matching between two matrices is not known in advance and investigate the quality 
of the solutions returned by MMOT-DC to solve the COOT and GW problems. This means that we will look at the COOT loss 
$\langle C, Q_s \otimes Q_f \rangle$, where the smaller the loss, the better when using both exact COOT and GW solvers
and our relaxation.

We generate two random matrices $X \in \mathbb R^{20 \times 3}$ and $Y \in \mathbb R^{30 \times 2}$, 
whose entries are drawn independently from the uniform distribution on the interval $[0,1)$. Then we calculate two corresponding 
squared Euclidean distance matrices of size $20$ and $30$. Their rows and columns are equipped with the discrete 
uniform distributions. In this case, \citep{Redko20} show that the COOT loss coincides with the GW distance, and the 
Block Coordinate Descent (BCD) algorithm used to approximate COOT is equivalent to the Frank-Wolfe algorithm \citep{Frank56} 
used to solve the GW distance.

We compare four solvers:
\begin{enumerate}
  \item The Frank-Wolfe algorithm to solve the GW distance (GW-FW).

  \item The projected gradient algorithm to solve the entropic GW distance \citep{Peyre16} (EGW-PGD). We choose the regularization 
  parameter from $\{0.0008, 0.0016, 0.0032, 0.0064, 0.0128, 0.0256 \}$ and pick the one which corresponds to smallest COOT loss.

  \item The Block Coordinate Descent algorithm to approximate the entropic COOT \citep{Redko20} 
  (EGW-BCD), where two additional KL divergences corresponding to two couplings are introduced. 
  Both regularization parameters are tuned from $\{0, 0.0005, 0.001, 0.005, 0.01, 0.05, 0.1, 0.5, 1 \}$, 
  where $0$ means that there is no regularization term for the corresponding coupling and we pick the pair whose COOT loss is the smallest.

  \item The algorithm \ref{algo:dc_MMOT} to solve the MMOT-DC. We tune 
  $\varepsilon \in \{1, 1.4, 1.8, 2.2, 2.6\}$ and we pick the one which corresponds to smallest COOT loss.
\end{enumerate}
For GW-FW and EGW-PGD, we use the implementation from the library \texttt{PythonOT} \citep{Flamary21}.

\begin{figure}[t]
	\centering
	\includegraphics[width=\textwidth,height=\textheight,keepaspectratio]{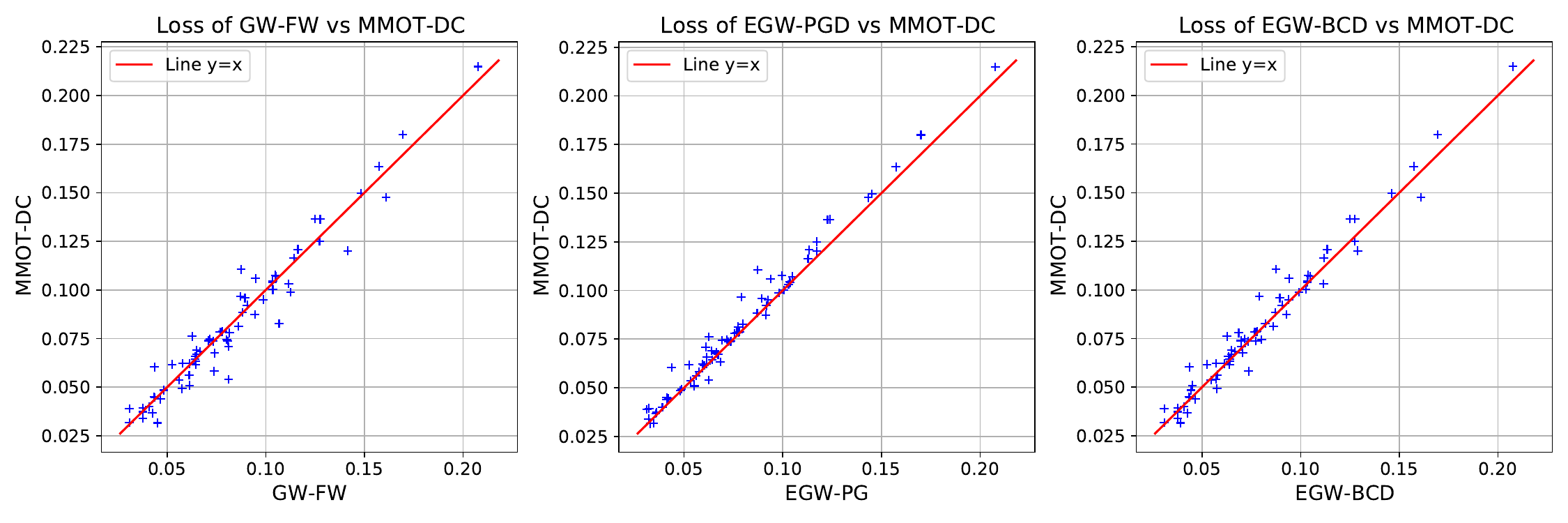}
	\caption{Scatter plots of MMOT-DC versus other solvers. In all three plots, the points tend to concentrate around the line $y=x$, 
  which indicates the comparable performance of MMOT-DC. On the other hand, the top-right plot shows the clear superiority of EGW-PGD.}
	\label{fig:gw}
\end{figure}

Given two random matrices, we record the COOT loss corresponding to the solution generated by each method. 
We simulate this process $70$ times and compare their overall performance. We can see in Table \ref{tab:gw} the average value and 
standard deviation and the comparison for the values of the loss between the different algorithms in Figure \ref{fig:gw}.
The performance is quite similar across methods with a  slight advantage for EGW-PGD. This is in itself a very
interesting result that has never been noted, to the best of our knowledge: the reason that the entropic version of GW can
provide better solution than solving the exact problem, may be due to the "convexification" of the problem, thanks to the entropic 
regularization. Our approach is also interestingly better than the exact GW-FW, which illustrates that the relaxation might help in 
finding better solutions despite the non-convexity of the problem.
%%%%%%%%%%%%%%%%%%%%%%%%%%%%%%%%%%%%%%%
\begin{table}[t]
  % \vskip 0.15in
  \begin{center}
    \begin{small}
      \begin{sc}
        \begin{tabular}{|c|c|c|c|}
          \hline
          GW-FW & EGW-PGD & EGW-BCD & MMOT-DC \\
          \hline
          0.0829 ($\pm$ 0.0354) & \textbf{0.0786 ($\pm$ 0.0347)} & 0.0804 ($\pm$ 0.0353) & 0.0822 ($\pm$ 0.0364) \\
          \hline
        \end{tabular}
      \end{sc}
    \end{small}
  \end{center}
  \caption{Average and standard deviation of COOT loss of the solvers. MMOT-DC is competitive to other solvers, 
  except for EGW-PGD and EGW-BCD. 
  \label{tab:gw}}
  % \vskip -0.1in
\end{table}

%%%%%%%%%%%%%%%%%%%%%%%%%%%%%%%%%%%%%%%
%%%%%%%%%%%%%%%%%%%%%%%%%%%%%%%%%%%%%%%

%%%%%%%%%%%%%%%%%%%%%%%%%%%%%%%%%%%%%%%%%%%%%%
\section{Discussion and conclusion}
%%%%%%%%%%%%%%%%%%%%%%%%%%%%%%%%%%%%%%%%%%%%%%
In this paper, we present a novel relaxation of the factorized MMOT problem called \textit{MMOT-DC}. More precisely, we replace the 
hard constraint on factorization constraint by a smooth regularization term. The resulting problem 
not only enjoys an interpolation property between MMOT and factorized MMOT, but also is a DC problem, 
which can be solved easily by the DC algorithm. We illustrate the use of MMOT-DC the via some simulated experiments and show that 
it is competitive with the existing popular solvers of COOT and GW distance. 
One limitation of the current DC algorithm is that, it is not scalable because
it requires storing a full-size tensor in the gradient step computation. Thus, future
work may focus on more efficiently designed algorithms, in terms of both time and memory footprint. 
Moreover, incorporating additional structure on the cost tensor may also be computationally and practically beneficial. 
From a theoretical viewpoint, it is also interesting to study the extension of MMOT-DC to the continuous setting, 
which can potentially allow us to further understand the connection between GW distance and COOT.

\paragraph{Acknowledgements.} The authors thank to Thibault Séjourné and Titouan Vayer for the fruitful discussion on the GW distance. 
The authors thank the anonymous reviewers for their careful proofreading and invaluable suggestions. This work is partially funded 
by the projects OATMIL ANR-17-CE23-0012, OTTOPIA ANR-20-CHIA-0030 and 3IA Côte d'Azur Investments ANR-19-P3IA-0002 of the 
French National Research Agency (ANR). This research was produced within the 
framework of Energy4Climate Interdisciplinary Center (E4C) of IP Paris and Ecole des Ponts ParisTech. This research was supported 
by the 3rd Programme d’Investissements d’Avenir ANR-18-EUR-0006-02. This action benefited from the support of the Chair 
"Challenging Technology for Responsible Energy" led by l’X – Ecole Polytechnique and the Fondation de l’Ecole Polytechnique, 
sponsored by TOTAL, and the Chair "Business Analytics for Future Banking" sponsored by NATIXIS.
%%%%%%%%%%%%%%%%%%%%%%%%%%%%%%%%%%%%%%%%%%%%%%%%%%%%%%%%%%%%

\bibliographystyle{plainnat}
\bibliography{neurips_2021.bib}
%%%%%%%%%%%%%%%%%%%%%%%%%%%%%%%%%%%%%%%%%%%%%%%%%%%%%%%%%%%%

\newpage
\appendix

%%%%%%%%%%%%%%%%%%%%%%%%%%%%%%%%%%%%%%%%%
\section{Appendix}

\paragraph{Derivation of the Sinkhorn algorithm in entropic MMOT.} The corresponding entropic dual problem of the primal problem 
\ref{MMOT_primal} reads
\begin{equation*}
  \sup_{f_n \in \mathbb R^{a_n}} \sum_{n=1}^N \langle f_n, \mu_n \rangle - 
  \varepsilon \sum_{i_1,...,i_N} \exp\Big( \frac{\sum_n (f_n)_{i_n} - C_{i_1,...,i_N}}{\varepsilon} \Big) + \varepsilon.
\end{equation*}
For each $n \in [N]$ and $i_n \in [a_n]$, the first order optimality condition reads
\begin{equation*}
  0 = (\mu_n)_{i_n} - \exp\big( \frac{(f_n)_{i_n}}{\varepsilon} \big) 
  \sum_{i_{-n}} \exp\Big( \frac{\sum_{j \neq n} (f_j)_{i_j} - C_{i_1,...,i_N}}{\varepsilon} \Big),
\end{equation*}
where, with some abuse of notation, we write $i_{-n} = (i_1, ..., i_{n-1}, i_{n+1}, ..., i_N)$. Or, equivalently
\begin{equation*}
  (f_n)_{i_n} = \varepsilon \log (\mu_n)_{i_n} - \varepsilon \log \sum_{i_{-n}}
  \exp\Big( \frac{\sum_{j \neq n} (f_j)_{i_j} - C_{i_1,...,i_N}}{\varepsilon} \Big),
\end{equation*}
or even more compact form
\begin{equation*}
  f_n = \varepsilon \log \mu_n - \varepsilon \log \sum_{i_{-n}}
  \exp\Big( \frac{\sum_{j \neq n} (f_j)_{i_j} - C_{\cdot, i_{-n}}}{\varepsilon} \Big).
\end{equation*}
Using the primal-dual relation, we obtain the minimiser of the primal problem \ref{MMOT_primal} by
\begin{equation*}
  P_{i_1,...,i_N} = \exp\Big( \frac{\sum_n (f_n)_{i_n} - C_{i_1,...,i_N}}{\varepsilon} \Big),
\end{equation*}
for $i_n \in [a_n]$, with $n \in [N]$.
%%%%%%%%%%%%%%%%%%%%%%%%%%%%%%%%%%%%%%%%%
Similar to the entropic OT, the Sinkhorn algorithm \ref{algo:dual_mmot} is also usually implemented in log-domain to avoid numerical instability.
\begin{algorithm}[H]
  \caption{Sinkhorn algorithm for the entropic MMOT problem \ref{MMOT_primal} from \citep{Benamou14}.}
  \textbf{Input.} Histograms $\mu_1,...,\mu_N$, hyperparameter $\varepsilon > 0$, cost tensor $C$ and 
  tuple of initial dual vectors $(f^{(0)}_1, ... f^{(0)}_N)$.

  \textbf{Output.} Optimal transport plan $P$ and tuple of dual vectors $(f_1, ... f_N)$ (optional).
  \begin{enumerate}
    \item While not converge: for $n = 1, ..., N$,
    \begin{equation*}
      \begin{split}
        f^{(t+1)}_n &= \varepsilon \log \mu_n - \varepsilon \log \sum_{i_{-n}} 
        \Big[ \exp\Big( \frac{\sum_{j < n} (f^{(t+1)}_j)_{i_j} + \sum_{j > n} (f^{(t)}_j)_{i_j} - 
        C_{\cdot, i_{-n}}}{\varepsilon} \Big) \Big].
      \end{split}
    \end{equation*}
    \item Return tensor $P$, where for $i_n \in [a_n]$, with $n \in [N]$,
    \begin{equation*}
      P_{i_1,...,i_N} = \exp\Big( \frac{\sum_n (f_n)_{i_n} - C_{i_1,...,i_N}}{\varepsilon} \Big).
    \end{equation*}
  \end{enumerate}
  \label{algo:dual_mmot}
\end{algorithm}
%%%%%%%%%%%%%%%%%%%%%%%%%%%%%%%%%%%%%%%%%

\paragraph{F-MMOT of two components (i.e. $M=2$) is a variation of low nonnegative rank OT.} 
For the sake of notational ease, we only consider the simplest case, where $N=4$ and $M=2$ with $\mathcal T_1 = (1,2)$ and 
$\mathcal T_2 = (3,4)$. However, the same argument still holds in the general case. First, we define three reshaping operations.
\begin{itemize}
  \item vectorization: concatenates rows of a matrix into a vector.
  \begin{equation*}
    \text{vec}: \mathbb R^{m \times n} \to \mathbb R^{m n},
  \end{equation*}
  where each element $A_{i,j}$ of the matrix $A \in \mathbb R^{m \times n}$ is mapped to a unique element $b_{(i-1)n + j}$ of the vector 
  $b \in \mathbb R^{m n}$, with $A_{i,j} = b_{(i-1)n + j}$, for $i = 1, ..., m$ and $j = 1, ...,n$. Conversely, 
  each element $b_k$ is mapped to a unique element $A_{k // n, n - k \% n}$, for every $k = 1, ..., mn$. 
  Here, $k // n$ is the quotient of the division of $k$ by $n$ and $k \% n$ is the 
  remainder of this division, i.e. if $k = q n + r$, with $0 \leq r < n$, then $k // n = q$ and $k \% n = r$.
  
  \item Matrization: transforms a $4$D tensor to a $2$D tensor (matrix) by vectorizing the first two and the last 
  two dimensions of the tensor.
  \begin{equation*}
    \text{mat}: \mathbb R^{n_1 \times n_2 \times n_3 \times n_4} \to \mathbb R^{(n_1 n_2) \times (n_3 n_4)},
  \end{equation*}
  where, similar to the vectorization, each element $P_{i,j,k,l}$ of the tensor 
  $P \in \mathbb R^{n_1 \times n_2 \times n_3 \times n_4}$ is mapped to the unique element $A_{(i-1)n_2 + j, (k-1)n_4 + l}$ of the 
  matrix $A \in \mathbb R^{(n_1 n_2) \times (n_3 n_4)}$, with $P_{i,j,k,l} = A_{(i-1)n_2 + j, (k-1)n_4 + l}$.

  \item Concatenation: stacks vertically two equal-column matrices.
  \begin{equation*}
    \begin{split}
      \text{con}_v: &\mathbb R^{m \times d} \times \mathbb R^{n \times d} \to \mathbb R^{(m+n) \times d} \\
      & \big( (u_1, ..., u_m),(v_1, ..., v_n) \big) \to (u_1, ..., u_m, v_1, ..., v_n)^T.
    \end{split}
  \end{equation*}
  Or, stacks horizontally two equal-row matrices
  \begin{equation*}
    \begin{split}
      \text{con}_h: &\mathbb R^{n \times p} \times \mathbb R^{n \times q} \to \mathbb R^{n \times (p+q)} \\
      & \big( (u_1, ..., u_p),(v_1, ..., v_q) \big) \to (u_1, ..., u_p, v_1, ..., v_q).
    \end{split}
  \end{equation*}
\end{itemize}
\begin{lemma} \label{vec_mat}
  For any $4$-D tensor $P \in \mathbb R^{n_1 \times n_2 \times n_3 \times n_4}$, denote $\pi$ its matrisation. We have,
  \begin{equation*}
    \text{vec} \Big( \sum_{k,l} P_{\cdot, \cdot, k, l} \Big) = \sum_{n=1}^{n_3 n_4} \pi_{\cdot, n} = \pi 1_{n_3 n_4},
  \end{equation*}
  where $1_n$ is the vector of ones in $\mathbb R^n$.
\end{lemma}
\paragraph{Proof of lemma \ref{vec_mat}.} For $(i,j) \in [n_1] \times [n_2]$, we have
\begin{equation*}
  \begin{split}
    \text{vec}\Big(\sum_{k,l} P_{\cdot,\cdot, k, l}\Big)_{(i-1)n_2 + j} &= \sum_{k,l} P_{i,j,k,l} \\
    &= \sum_{k,l} \pi_{(i-1) n_2 + j, (k-1) n_4 + l} \\
    &= \sum_{n=1}^{n_3 n_4} \pi_{(i-1) n_2 + j, n}.
  \end{split}
\end{equation*}
The result then follows. \qed

Now, let $(e_i)_{i=1}^{n_1 n_2}$ be the standard basis vectors of $\mathbb R^{(n_1 n_2)}$, i.e. $(e_i)_k = 1_{\{i=k\}}$. 
For each $P \in U(\mu)$, denote $\pi$ its matrisation, then by lemma \ref{vec_mat}, we have, for $i \in [n_1]$,
\begin{equation*}
  (\mu_1)_i = \sum_j \sum_{k,l} P_{i,j,k,l} = \sum_{j = 1}^{n_2} \sum_{n=1}^{n_3 n_4} \pi_{(i-1) n_2 + j, n},
\end{equation*}
which can be recast in matrix form as
\begin{equation*}
  A_1^T \pi 1_{n_3 n_4} = \mu_1
\end{equation*}
where the matrix $A_1 = \text{con}_{h}(v_1, ..., v_{n_1}) \in \mathbb R^{(n_1 n_2) \times n_1}$, with 
$v_i \in \mathbb R^{(n_1 n_2)}$, where $v_i = \sum_{j=(i-1)n_2 + 1}^{i n_2} e_j$, with $i \in [n_1]$. Similarly, 
$A_2 \pi 1_{n_3 n_4} = \mu_2$, where the matrix 
$A_2 = \text{con}_h(I_{n_2}, ..., I_{n_2}) \in \mathbb R^{n_2 \times (n_1 n_2)}$, where $I_n \in \mathbb R^{n \times n}$ 
is the identity matrix. Both conditions can be compactly written as
\begin{equation*}
  A_{12}^T \pi 1_{n_3 n_4} = \mu_{12},
\end{equation*}
where the matrix $A_{12} = \text{con}_h(A_1, A_2^T) \in \mathbb R^{(n_1 n_2) \times (n_1+n_2)}$ and 
$\mu_{12} = \text{con}_v(\mu_1, \mu_2) \in \mathbb R^{(n_1 + n_2)}$. Note that $\mu_{12}$ is not a probability 
because its mass is $2$. The matrix $A_{12}$ has exactly $2n_1n_2$ ones and the rest are zeros. Similarly, 
for $A_{34}$ and $\mu_{34}$ defined in the same way as $A_{12}$ and 
$\mu_{12}$, respectively, we establish the equality $A_{34}^T \pi^T 1_{n_1 n_2} = \mu_{34}$. 
As a side remark, both matrices $A_{12}^T$ and $A_{34}^T$ are \textit{totally unimodular}, i.e. every square submatrix has determinant 
$-1, 0$, or $1$.
\begin{figure}[ht]
  \centering
  \includegraphics[width=0.25\textwidth,height=0.25\textheight,keepaspectratio]{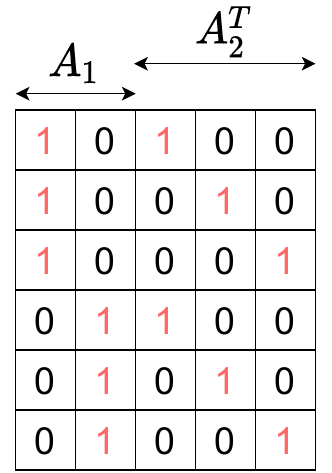}
  \caption{An example of the matrix $A_{12}$ when $n_1=2$ and $n_2=3$.}
  \label{fig:matrix}
\end{figure}

To handle the factorization constraint, first we recall the following concept.
\begin{definition}
  Given a nonnegative matrix $A$, we define its nonnegative rank by
  \begin{equation*}
    \text{rank}_{+}(A):= \min \big\{ r \geq 1: A = \sum_{i=1}^r M_i, \text{ where } \text{rank}(M_i) = 1, M_i \geq 0, \forall i \big\}.
  \end{equation*}
  By convention, zero matrix has zero (thus nonnegative) rank.
\end{definition}
So, the constraint $P = P_1 \otimes P_2$ is equivalent to $\text{mat}(P) = \text{vec}(P_1) \text{vec}(P_2)^T$. 
By lemma 2.1 in \citep{Joel93}, $\text{rank}_+(A) = 1$ if and only if there exist two nonnegative vectors $u,v$ such that 
$A = u v^T$. Thus, the factorization constraint is equivalent to $\text{rank}_+\big( \text{mat}(P) \big) = 1$.

Denote $L= \text{mat}(C)$ and $M = n_1 n_2, N = n_3 n_4$. Now, the problem \ref{factor_mmot} can be rewritten as
\begin{equation*}
  \begin{split}
    \min_{Q \in \mathbb R^{M \times N}_{\geq 0}} &\langle L, Q \rangle \\
    \text{ such that } & A_{12}^T Q 1_N = \mu_{12} \\
    &A_{34}^T Q^T 1_M = \mu_{34} \\
    &\text{rank}_{+}(Q) = 1,
  \end{split}
\end{equation*}
which is a variation of the low nonnegative rank OT problem studied in \citep{Meyer21a}. \qed

%%%%%%%%%%%%%%%%%%%%%%%%%%%%%%%%%%%%%%%%%
\paragraph{Proof of proposition \ref{MMOT_dc_prop}.} The inequality 
$\text{MMOT}(\mu) \leq \text{MMOT-DC}_{\varepsilon}(\mathcal T, \mu)$ follows from 
the positivity of the KL divergence. On the other hand,
\begin{equation*}
  \text{F-MMOT}(\mathcal T, \mu) = 
  \inf_{P \in U_{\mathcal T}} \langle C, P \rangle + \varepsilon \text{KL}(P \vert P_{\# \mathcal T}),
\end{equation*}
because $\text{KL}(P \vert P_{\# \mathcal T}) = 0$, for every $P \in U_{\mathcal T}$. As 
$U_{\mathcal T} \subset U(\mu)$, we have 
$\text{MMOT-DC}_{\varepsilon}(\mathcal T, \mu) \leq \text{F-MMOT}(\mathcal T, \mu)$.

Now, if $\text{F-MMOT}(\mathcal T, \mu) = 0$, then $\text{MMOT-DC}_{\varepsilon}(\mathcal T, \mu) = 0$. Conversely, 
if $\text{MMOT-DC}_{\varepsilon}(\mathcal T, \mu) = 0$, for $\varepsilon > 0$, 
then there exists $P^* \in U(\mu)$ such that $\langle C, P^* \rangle = 0$ and $P^* = P^*_{\# \mathcal T}$. 
Thus $\langle C, P^*_{\# \mathcal T} \rangle = 0$, which means $\text{F-MMOT}(\mathcal T, \mu) = 0$. \qed

%%%%%%%%%%%%%%%%%%%%%%%%%%%%%%%%%%%%%%%%%
\paragraph{Proof of proposition \ref{interpolation_prop}.}
The function $\varepsilon \to \text{MMOT-DC}_{\varepsilon}(\mathcal T, \mu)$ is increasing on $\mathbb R_{\geq 0}$ and bounded, 
thus admits a finite limit $L \leq \text{F-MMOT}(\mathcal T, \mu)$, when $\varepsilon \to \infty$, and a finite limit 
$l \geq \text{MMOT}(\mu)$, when $\varepsilon \to 0$.

Let $P_{\varepsilon}$ be a solution of the problem $\text{MMOT-DC}_{\varepsilon}(\mathcal T, \mu)$. 
As $U(\mu)$ is compact, when either $\varepsilon \to 0$ or $\varepsilon \to \infty$, 
one can extract a converging subsequence (after reindexing) 
$(P_{\varepsilon_k})_k \to \widetilde{P} \in U(\mu)$, 
when either $\varepsilon_k \to 0$ or $\varepsilon_k \to \infty$. 
Thus, the convergence of the marginal distributions is also guaranteed, i.e 
$(P_{\varepsilon_k})_{\# \mathcal T_m} \to \widetilde{P}_{\# \mathcal T_m} \in U_{\mathcal T_m}$, for every $m \in [M]$, 
which implies that $P_{\varepsilon_k} - (P_{\varepsilon_k})_{\# \mathcal T} \to \widetilde{P} - \widetilde{P}_{\# \mathcal T}$.

When $\varepsilon \to 0$, let $P^{*}$ be a solution of the problem $\text{MMOT}(\mu)$. Then,
\begin{equation*}
  \langle C, P^* \rangle \leq \langle C, P_{\varepsilon} \rangle + 
  \varepsilon \text{KL}(P_{\varepsilon} \vert (P_{\varepsilon})_{\# \mathcal T}) \leq 
  \langle C, P^{*} \rangle + \varepsilon \text{KL}(P^* \vert P^*_{\# \mathcal T}).
\end{equation*}
By the sandwich theorem, when $\varepsilon \to 0$, we have 
$\text{MMOT-DC}_{\varepsilon}(\mathcal T, \mu) \to \langle C, P^* \rangle = \text{MMOT}(\mu)$. 
Furthermore, as
\begin{equation*}
  0 \leq \langle C, P_{\varepsilon_k} \rangle - \langle C, P^* \rangle 
  \leq \varepsilon_k \text{KL}(P^* \vert P^*_{\# \mathcal T}),
\end{equation*}
when $\varepsilon_k \to 0$, it follows that $\langle C, \widetilde{P} \rangle = \langle C, P^* \rangle$. 
So $\widetilde{P}$ is a solution of the problem $\text{MMOT}(\mu)$. We conclude that 
any cluster point of the sequence of minimisers of $\text{MMOT-DC}_{\varepsilon}(\mathcal T, \mu)$ when 
$\varepsilon \to 0$ is a minimiser of $\text{MMOT}(\mu)$. As a byproduct, since       
\begin{equation*}
  \text{KL}(P^* \vert P^*_{\# \mathcal T}) - \text{KL}(P_{\varepsilon_k} \vert (P_{\varepsilon_k})_{\# \mathcal T}) \geq
  \frac{\langle C, P_{\varepsilon_k} \rangle - \langle C, P^* \rangle}{\varepsilon_k} \geq 0,
\end{equation*}
we also deduce that $\text{KL}(\widetilde{P} \vert \widetilde{P}_{\# \mathcal T}) \leq \text{KL}(P^* \vert P^*_{\# \mathcal T})$ 
(so the cluster point $\widetilde{P}$ has minimal "mutual information").

On the other hand, when $\varepsilon \to \infty$, for $\mu^{\otimes N} = \mu_1 \otimes ... \otimes \mu_N$, one has
\begin{equation*}
  \langle C, \mu^{\otimes N} \rangle + \varepsilon \times 0 \geq \langle C, P_{\varepsilon} \rangle + 
  \varepsilon \text{KL}(P_{\varepsilon} \vert (P_{\varepsilon})_{\# \mathcal T})
  \geq \varepsilon \text{KL}(P_{\varepsilon} \vert (P_{\varepsilon})_{\# \mathcal T}).
\end{equation*}
Thus,
\begin{equation*}
  0 \leq \text{KL}(P_{\varepsilon} \vert (P_{\varepsilon})_{\# \mathcal T}) \leq 
  \frac{1}{\varepsilon} \langle C, \mu^{\otimes N} \rangle \to 0, \text{ when } \varepsilon \to \infty,
\end{equation*}
which means $\text{KL}(P_{\varepsilon} \vert (P_{\varepsilon})_{\# \mathcal T}) \to 0$, when $\varepsilon \to \infty$. In particular, 
when $\varepsilon_k \to \infty$, we have $\text{KL}(P_{\varepsilon_k} \vert (P_{\varepsilon_k})_{\# \mathcal T}) \to 0$. We deduce that 
$\text{KL}(\widetilde{P} \vert \widetilde{P}_{\# \mathcal T}) = 0$, which implies $\widetilde{P} = \widetilde{P}_{\# \mathcal T}$.

Now, as $\text{MMOT-DC}_{\varepsilon}(\mathcal T, \mu) \geq \langle C, P_{\varepsilon} \rangle$, 
when $\varepsilon \to \infty$, we have $L \geq \langle C, \widetilde{P} \rangle = 
\langle C, \widetilde{P}_{\# \mathcal T} \rangle \geq \text{F-MMOT}(\mathcal T, \mu)$. 
Thus $L = \langle C, \widetilde{P} \rangle = \text{F-MMOT}(\mathcal T, \mu)$, i.e.
$\text{MMOT-DC}_{\varepsilon}(\mathcal T, \mu) \to \text{F-MMOT}(\mathcal T, \mu)$ when $\varepsilon \to \infty$. In this case, 
we also have that any cluster point of the sequence of minimisers of $\text{MMOT-DC}_{\varepsilon}(\mathcal T, \mu)$ 
is a minimiser of $\text{F-MMOT}(\mathcal T, \mu)$. \qed

%%%%%%%%%%%%%%%%%%%%%%%%%%%%%%%%%%%%%%%%%
\paragraph{Proof of corollary \ref{kernel_gw_coot}.} In this proof, we write $C: = L(C_x,C_y)$, for notational convenience. 
In the setting of GW distance, we have $N=4$ and $M = 2$ with $\mathcal T_1 = (1,2)$ and $\mathcal T_2 = (3,4)$. 
Given a solution $P_{\varepsilon}$ of the problem \ref{relax_mmot}, we also write 
$P_{\varepsilon, i} := (P_{\varepsilon})_{\# \mathcal T_i}$, for short. Now, for $i = 1,2$, 
let $Q_i \in U( P_{\varepsilon, i}, P_{\varepsilon, i}) \subset U(\mu)$. The optimality of $P_{\varepsilon}$ 
implies that
\begin{equation*}
    \langle C, P_{\varepsilon} \rangle + \varepsilon \big[ H(P_{\varepsilon}) - H(P_{\varepsilon, 1}) - 
    H(P_{\varepsilon, 2}) \big] \leq 
    \langle C, Q_i \rangle + \varepsilon \big[ H(Q_i) - 2 H(P_{\varepsilon, i}) \big].
\end{equation*}
Thus,
\begin{equation*}
    2 \big( \langle C, P_{\varepsilon} \rangle + \varepsilon H(P_{\varepsilon}) \big) \leq 
    \sum_{i=1}^2 \langle C, Q_i \rangle + \varepsilon H(Q_i).
\end{equation*}
As this is true for every $Q_i \in U(P_{\varepsilon, i}, P_{\varepsilon, i})$, we have
\begin{equation*}
    \begin{split}
        \frac{1}{2} \sum_{i=1}^2 
        \text{OT}_{\varepsilon}(P_{\varepsilon, i}, P_{\varepsilon, i})
        &= \frac{1}{2} \sum_{i=1}^2 \inf_{Q_i \in U(P_{\varepsilon, i}, P_{\varepsilon, i})} 
        \langle C, Q_i \rangle + \varepsilon H(Q_i) \\
        &\geq \langle C, P_{\varepsilon} \rangle + \varepsilon H(P_{\varepsilon}) \\
        &\geq \inf_{P \in U(P_{\varepsilon, 1}, P_{\varepsilon, 2})} 
        \langle C, P \rangle + \varepsilon H(P) \\
        &= \text{OT}_{\varepsilon}(P_{\varepsilon, 1}, P_{\varepsilon, 2}).
    \end{split}
\end{equation*}
The second inequality holds because $P_{\varepsilon} \in U(P_{\varepsilon, 1}, P_{\varepsilon, 2})$. Thus,
\begin{equation} \label{sinkhorn_div}
  \text{OT}_{\varepsilon}(P_{\varepsilon, 1}, P_{\varepsilon, 2}) - 
  \frac{1}{2} \sum_{i=1}^2 \text{OT}_{\varepsilon}(P_{\varepsilon, i}, P_{\varepsilon, i}) \leq 0.
\end{equation}
The left-hand side of the inequality \ref{sinkhorn_div} is nothing but the Sinkhorn divergence between $P_{\varepsilon, 1}$ and 
$P_{\varepsilon, 2}$ \citep{Ramdas17}. As the kernel $C$ is conditionally negative definite if and only if for every 
$\varepsilon > 0$, the kernel $e^{-C / \varepsilon}$ is positive definite \citep{Schoenberg38}, by proposition 5 in \citep{Janati20}, 
the inequality in \ref{sinkhorn_div} becomes an equality. As a consequence, for $i=1,2$, if 
$Q_{\varepsilon, i} \in U( P_{\varepsilon, i}, P_{\varepsilon, i})$ is the (unique) optimal plan of the entropic OT problem 
$\text{OT}_{\varepsilon}(P_{\varepsilon, i}, P_{\varepsilon, i})$, then we must have
\begin{equation*}
  \langle C, P_{\varepsilon} \rangle + \varepsilon \big[ H(P_{\varepsilon}) - H(P_{\varepsilon, 1}) - 
  H(P_{\varepsilon, 2}) \big] =  
  \langle C, Q_{\varepsilon, i} \rangle + \varepsilon \big[ H(Q_{\varepsilon, i}) - 2 H(P_{\varepsilon, i}) \big],
\end{equation*}
or equivalently, $Q_{\varepsilon, 1}$ and $Q_{\varepsilon, 2}$ are also solutions of the problem \ref{relax_mmot}.

Now, by proposition \ref{interpolation_prop}, when $\varepsilon \to \infty$, a cluster point 
$P^* = P^*_{\# \mathcal T_1} \otimes P^*_{\# \mathcal T_2}$ of the sequence of minimisers $(P_{\varepsilon})_{\varepsilon}$ induces a solution 
$(P^*_{\# \mathcal T_1}, P^*_{\# \mathcal T_2})$ of the COOT problem. In particular, $P^*_{\# \mathcal T_i}$ is a cluster point of 
$(P_{\varepsilon, i})_{\varepsilon}$ and there exists a cluster point $Q^*_i$ of 
$(Q_{\varepsilon, i})_{\varepsilon}$ in $U(P^*_{\# \mathcal T_i}, P^*_{\# \mathcal T_i})$, for $i=1,2$. But still by proposition 
\ref{interpolation_prop}, we also have that $Q^*_i = (Q^*_i)_{\# \mathcal T_1} \otimes (Q^*_i)_{\# \mathcal T_2}$. Thus, 
$Q^*_i = P^*_{\# \mathcal T_i} \otimes P^*_{\# \mathcal T_i}$ 
and the solution $(P^*_{\# \mathcal T_1}, P^*_{\# \mathcal T_2})$ of the COOT problem satisfies: 
$\langle C, P^*_{\# \mathcal T_1} \otimes P^*_{\# \mathcal T_2} \rangle = 
\langle C, P^*_{\# \mathcal T_1} \otimes P^*_{\# \mathcal T_1} \rangle = 
\langle C, P^*_{\# \mathcal T_2} \otimes P^*_{\# \mathcal T_2} \rangle$. The equality between GW distance and COOT then follows, 
and $P^*_{\# \mathcal T_1}$ and $P^*_{\# \mathcal T_2}$ are two solutions of the GW problem.

If furthermore, the kernel $C$ induces a strictly positive definite kernel, then by proposition 5 in \citep{Janati20}, 
we deduce that $P_{\varepsilon, 1} = P_{\varepsilon, 2}$. One can also use the following reasoning: 
in the finite setting, a strictly positive definite kernel is necessarily universal (see for example section 2.3 in 
\citep{Borgwardt06}), and the kernel $C$ defined on $(\mathcal X \times \mathcal Y)^2$ is necessarily a (symmetric) 
Lipschitz function with respect to both inputs. So, the Sinkhorn divergence vanishes if and only if 
$P_{\varepsilon, 1} = P_{\varepsilon, 2}$ \citep{Feydy19}. From either reasoning, we conclude that 
$P^*_{\# \mathcal T_1} = P^*_{\# \mathcal T_2}$. \qed

\paragraph{An empirical variation.} Intuitively, for sufficiently large $\varepsilon$, the minimisation of the KL divergence is prioritised 
over the linear term in the objective function of the MMOT-DC problem, which implies that the optimal tensor $P^*$ is "close" to its 
corresponding tensor product $P^*_{\# \mathcal T}$. So, instead of calculating the gradient at $P$, one may calculate at 
$P_{\# \mathcal T}$. In this case, the gradient reads
\begin{equation*}
  \begin{split}
    \sum_{m=1}^M \nabla_P H_m(P_{\# \mathcal T}) = 
    \big[ \log P_{\# \mathcal T_1} + P_{\# \mathcal T_1} \big] \oplus ... \oplus \big[ \log P_{\# \mathcal T_M} + P_{\# \mathcal T_M} \big],
  \end{split}
\end{equation*}
where $\oplus$ represents the tensor sum operator between two arbitrary-size tensors: $(A \oplus B)_{i,j}:= A_i + B_j$, where with some 
abuse of notation, $i$ or $j$ can be understood as a tuple of indices. Thus, we avoid storing the $N$-D gradient tensor (as in the 
algorithm \ref{algo:dc_MMOT}) and only need to store $M$ smaller-size tensors. Not only saving the memory, 
this variation also seems to be empirically competitive with the original algorithm \ref{algo:dc_MMOT}, if not sometimes better, 
in terms of COOT loss. The underlying reason might be related to the approximate DCA scheme \citep{Thanh15}, where one replaces both 
steps in each DC iteration by their approximation. We leave the formal theoretical justification of this variation to the future work.
We call this variation \textit{MMOT-DC-v1} and use the same setup as in the experiment \ref{expe:2}.
\begin{figure}[ht]
  \centering
  \includegraphics[width=\textwidth,height=\textheight,keepaspectratio]{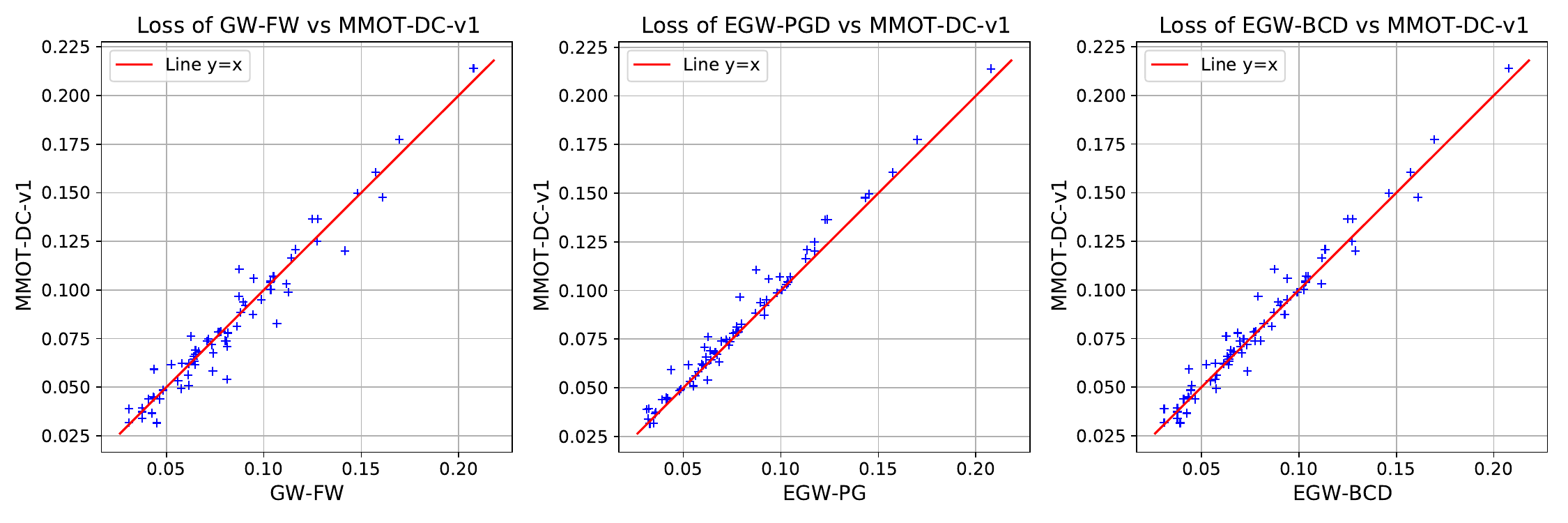}
  \caption{Scatter plots of MMOT-DC-v1 versus other solvers. In all three plots, the points tend to concentrate around the line $y=x$, 
  which indicates the comparable performance of MMOT-DC-v1. On the other hand, the top-right plot shows the clear superiority of EGW-PGD.}
  \label{fig:coot_mmot_new}
\end{figure}

\begin{table}[H]
  \label{tab:coot_new}
  % \vskip 0.15in
  \begin{center}
    \begin{small}
      \begin{sc}
        \begin{tabular}{|c|c|}
          \hline
          MMOT-DC & MMOT-DC-v1 \\
          \hline
          0.0822 ($\pm$ 0.0364) & 0.0820 ($\pm$ 0.0361) \\
          \hline
        \end{tabular}
      \end{sc}
    \end{small}
  \end{center}
  \caption{Average and standard deviation of COOT loss of MMOT-DC and MMOT-DC-v1. The performance of the two algorithms is 
  very similar.}
  % \vskip -0.1in
\end{table}

\end{document}